\pdfoutput=1
\documentclass[11pt]{article}

\usepackage[final]{coling}

\usepackage{times}
\usepackage{latexsym}

\usepackage[T1]{fontenc}
\usepackage[utf8]{inputenc}

\usepackage{microtype}

\usepackage{inconsolata}

\usepackage{graphicx}

\usepackage{enumitem}
\usepackage{amsmath}
\usepackage{multirow}
\usepackage{booktabs}
\usepackage{array}
\usepackage{wrapfig}

\title{LLaVA-RE: Binary Image-Text Relevancy Evaluation with Multimodal Large Language Model}

\author{Tao Sun\thanks{Work done during internship at Amazon.} \\
  Stony Brook University \\
  \texttt{tao@cs.stonybrook.edu} \\\And
  Oliver Liu, JinJin Li, Lan Ma \\
  Amazon \\
  \texttt{\{olivlius,jinjinli,mamlm\}@amazon.com} \\}

\begin{document}
\maketitle
\begin{abstract}

Multimodal generative AI usually involves generating image or text responses given inputs in another modality. The evaluation of image-text relevancy is essential for measuring response quality or ranking candidate responses. In particular, binary relevancy evaluation, \textit{i.e.}, ``Relevant'' \textit{vs.} ``Not Relevant'', is a fundamental problem. However, this is a challenging task considering that texts have diverse formats and the definition of relevancy varies in different scenarios. We find that Multimodal Large Language Models (MLLMs) are an ideal choice to build such evaluators, as they can flexibly handle complex text formats and take in additional task information. In this paper, we present LLaVA-RE, a first attempt for binary image-text relevancy evaluation with MLLM. It follows the LLaVA architecture and adopts detailed task instructions and multimodal in-context samples. In addition, we propose a novel binary relevancy data set that covers various tasks. Experimental results validate the effectiveness of our framework.

%When evaluated on unseen and fine-grained relevancy tasks, our model obtains superior accuracies compared than state-of-the-art MLLMs. 

\end{abstract}

\section{Introduction}
Multimodal generative AI such as GPT-4V~\cite{achiam2023gpt}, Gemini~\cite{team2023gemini}, and Stable Diffusion~\cite{rombach2022high} has shown remarkable ability to generate image or text responses. A typical scenario is an AI assistant where \textit{agent} responses to \textit{user} instructions during a conversation. For example, \textit{user} inputs a textual query, and \textit{agent} returns an image that is generated or retrieved from some database. To measure response quality or rank candidate responses, an essential component is evaluating the relevancy between text and image. However, this is not an easy task. The texts can have diverse formats such as a long description, a multi-turn conversation, or a structured document digest. Such complex texts usually contain rich information, and the definition of relevancy varies in different scenarios. It requires to specify attributes that lead to a `relevant' image. For example, a multi-turn conversation and an image may talk about the same product but have some controversial details, such as color or size; when describing fine-grained bird species, one image can match common attributes of the bird genus but not specie-wise details. In both cases, the image can be labeled either as `relevant' or `not relevant', depending on the particular goal. Traditional retrieval models~\cite{frome2013devise,lee2018stacked,qu2021dynamic} rely on image and text embeddings. They are not suitable for this complex evaluation task with long texts. Methods like CLIP~\cite{radford2021learning} and BLIP~\cite{li2022blip} fall apart for long and ambiguous texts. 

These challenges motivate us to build an effective relevancy evaluation model for complex image-text pairs. We focus on binary image-text relevancy, \textit{i.e.}, ``Relevant'' \textit{vs.} ``Not Relevant''. Although it is possible to add intermediate relevancy labels such as ``Somewhat Relevant'', binary relevance labels are more common in practical usage and it enforces evaluators to make less ambiguous labeling.

Multimodal Large Langauge Models (MLLMs) such as LLaVA~\cite{liu2024visual} are an ideal choice for the above-mentioned purposes. Compared with traditional models that rely on similarity scores between image and text embeddings~\cite{wang2018learning}, MLLMs exhibit much more flexibility. As MLLMs are pre-trained on huge image-text corpus, they can easily handle diverse text formats. Besides, additional task information such as the relevancy definition or demonstration examples can be readily integrated into model inputs. However, even with contextual information, a direct extension of state-of-the-art MLLMs does not perform effectively on relevancy tasks.

In this paper, we present \textbf{L}arge \textbf{L}anguage \textbf{a}nd \textbf{V}ision \textbf{A}ssistant for binary image-text \textbf{R}elevancy \textbf{E}valuation (LLaVA-RE), a first attempt for relevancy evaluation with MLLM. Our model builds upon the LLaVA 1.5 architecture~\cite{liu2024improved}, which shows excellent performances among open-sourced MLLMs and can be easily extended owing to its light-weight design. To handle ambiguity in relevancy, we adopt detailed task instructions. Furthermore, we leverage multimodal in-context-learning~\cite{doveh2024towards} to include few-shot demonstration examples. These designs empower LLaVA-RE to generalize to unseen relevancy tasks and achieve more accurate predictions. Since there are no publicly available datasets focusing on complex image-text relevancy, we propose a novel binary relevancy dataset covering diverse tasks. For each task, a strategy to sample positive and negative image-text pairs is delicately designed. We train our model on the curated datasets and evaluate on unseen and fine-grained relevancy tasks. 

%Compared with state-of-the-art LLaVA models, ours obtain better accuracies. Further, on fine-grained tasks, LLaVA-RE can even match a powerful Idefics2 model that is pre-trained on far more data.

We summarize the contributions as follows:
\begin{itemize}[itemsep=0pt,topsep=0pt,parsep=0pt]
	\item To the best of our knowledge, LLaVA-RE is the first work to build MLLM for binary image-text relevancy evaluation. 
	\item We create a novel binary relevancy dataset covering diverse tasks, where positive and negative image-text pairs are delicately sampled.
	\item Experimental results validate the effectiveness of our framework over the vanilla LLaVA 1.5 by incorporating novel designs of task instructions and multimodal in-context learning.
    %With novel designs of task instructions and multimodal in-context samples, LLaVA-RE achieves superior performances on various evaluation tasks.
\end{itemize}

\begin{figure*}[!t]
    \centering
    \includegraphics[width=1.0\textwidth]{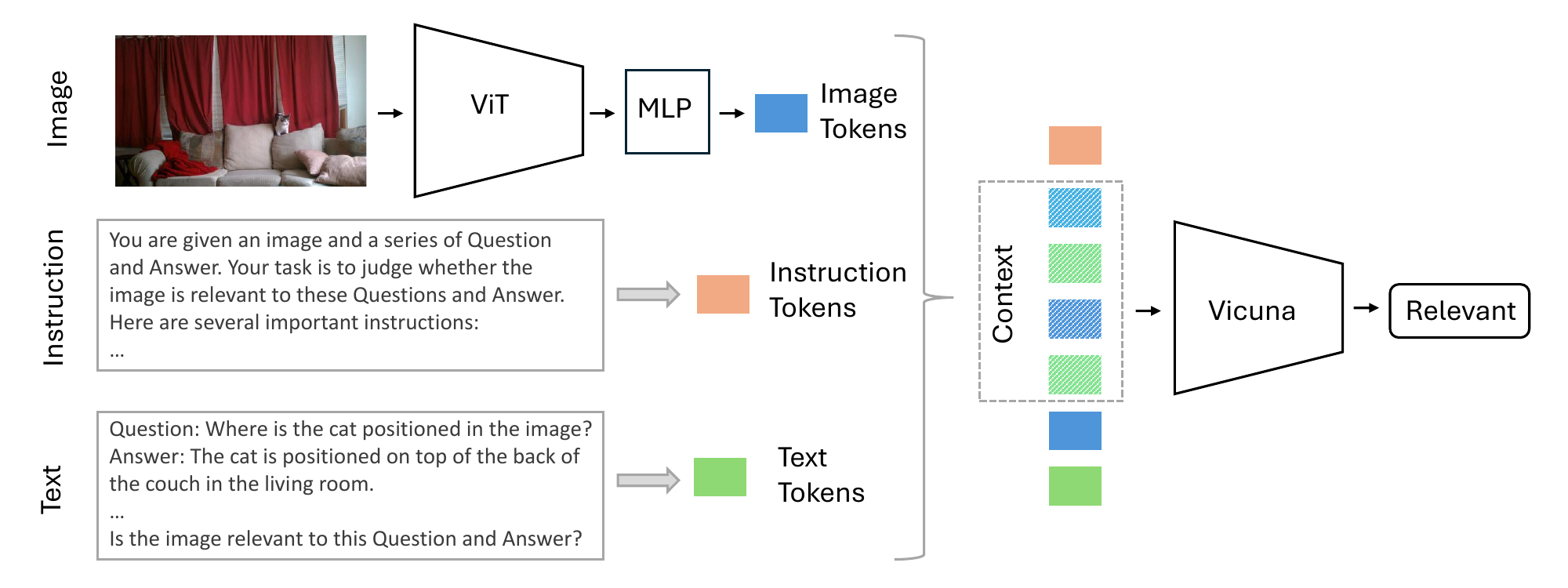}
    \caption{Framework of LLaVA-RE model. We use ViT and Vicuna as the image and text encoder, respectively. Context samples are selected from the same relevancy evaluation task.}
    \label{fig:app:framework}
\end{figure*}

\section{Related Work}
\paragraph{Image-Text Retrieval} is a common task that retrieves the most related image or text giving the counterpart~\cite{cao2022image}. Traditional methods~\cite{frome2013devise,lee2018stacked,qu2021dynamic} built visual semantic embeddings and model dense cross-modal interactions to get similarity scores. CLIP~\cite{radford2021learning} is a pioneering work that aligns image and text modalities via contrastive learning on abundant imaga-text pairs. This is later improved with bootstrapping~\cite{li2022blip} and Query Transformer~\cite{li2023blip}. InternVL~\cite{chen2024internvl} scaled up the vision foundation model and progressively aligns it with LLM. Although these works aim to match image and text, their texts are often short image captions. In contrast, we tackle relevancy evaluation tasks that involve significantly longer texts, greater ambiguity, and more complex formats.

\paragraph{MLLM and Binary VQA.}
Recently, numerous MLLM models have been introduced~\cite{achiam2023gpt,laurenccon2024matters}, with the LLaVA family~\cite{liu2024visual} being the most closely related to our work. LLaVA model connected a pretrained vision encoder and an LLM with a linear layer, and trains on visual instruction-following data generated by GPT-4. LLaVA 1.5 enhances performance with an MLP projector and academic-task VQA datasets~\cite{liu2024improved}, while~\citet{liu2024llavanext} introduce dynamic image resolution and stronger LLM backbones. MLLM models achieve impressive performances on diverse Visual Question Answering (VQA) tasks, and binary VQA that has a ``yes/no'' answer is an important subset. However, these existing binary VQA questions hae simple forms and it is unclear how MLLMs generalize to the challenging text-image relevancy tasks studied in this paper.

 % covering object recognition~\cite{goyal2017making}, text reasoning~\cite{singh2019towards}, scientific questions~\cite{lu2022learn}, \textit{etc}

\paragraph{Multimodal In-Context-Learning} utilizes multimodal context to improve model inference. \citet{li2023mimic} construct an interleaved multi-modal ICL dataset and train a Flamingo-based model to demonstrate ICL capability. \citet{zhao2023mmicl} introduce a novel context scheme that incorporates an additional image declaration section and includes image proxy tokens to enhance model’s ICL ability. \citet{doveh2024towards} extend LLaVA with ICL capability by tuning on few-shot instruction data. Despite these innovations, the effectiveness to incorporate ICL in binary relevancy tasks is under-explored. Our work finds current multimodal ICL solutions struggling to adapt effectively to this specific relevancy evaluation task.

\section{Approach}

\subsection{Binary Relevancy Evaluation Formulation}
Given a pair of image $I$ and text $T$, we want to evaluate whether they are relevant or not. Formally, a relevancy evaluator $\mathcal{M}$ maps $(I,T)$ into a binary label $r\in\{\rm{``Relevant"}, \rm{``Not\ Relevant"}\}$. Usually, this is not a well-defined task as the meaning of relevancy depends on specific scenarios. We assume that there exists an additional task instruction $S$, which is a paragraph of natural language describing the data and clarifying the relevancy definition. Meanwhile, there could be a few demonstration examples $\{(I_i,T_i,r_i)\}$ from the same task. Binary relevancy evaluation can be formulated as follows:
\begin{equation}
    r=\mathcal{M}(I, T; S, \{(I_i,T_i,r_i)\})
\end{equation}

\begin{table}[!t]
\small
    \centering
    \begin{tabular}{p{1.6cm}p{0.45cm}p{0.45cm}p{3.5cm}}
    \toprule
         Task &  Train & Test & Text format  \\
         \midrule
         llava & 10$k$ & 6$k$  & Conversations  \\
         wiki & 20$k$ & 300 & Plain paragraph  \\
         recipe & 12$k$ & 1$k$ & Ingredients description  \\
         %news & 20$k$ & 2$k$ & Structured news  \\
         textvqa & 33$k$ & 1$k$ & Question, answer, reasoning  \\
         tdiuc & 7$k$ & 300 & Question, answer, reasoning \\ 
         chartqa & -- & 1$k$ & Question, answer, reasoning \\
         infographics & -- & 1$k$ & Question, answer, reasoning \\
         fine-grained & -- & 6$k$ & Category description \\
    \bottomrule
    \end{tabular}
    \caption{List of created binary relevancy datasets.}
    \label{tab:list_of_datasets}
\end{table}

\subsection{LLaVA-RE}

In this paper, we present \textbf{L}arge \textbf{L}anguage \textbf{a}nd \textbf{V}ision \textbf{A}ssistant for binary image-text \textbf{R}elevancy \textbf{E}valuation (LLaVA-RE). 
It is built upon the LLaVA 1.5 architecture~\cite{liu2024improved}, which uses a conversation data format and can readily integrate task instructions and demonstration text-image examples. One data sample is as follows:
\begin{equation}
  \begin{aligned}
    &\texttt{Human}: S \\
    &\texttt{Human}: I_1, T_1 \quad \texttt{Assistant}: r_1  \\
    & ...\\
    &\texttt{Human}: I_C, T_C \quad \texttt{Assistant}: r_C  \\
    &\texttt{Human}: I, T \quad \texttt{Assistant}: r 
  \end{aligned}
\end{equation}
where $C$ is the number of demonstration examples.

The model training includes two stages: first, we train the image projector using the same methodology as LLaVA 1.5~\cite{liu2024improved}; second, we train the language backbone with multimodal ICL instruction tuning~\cite{doveh2024towards} using binary relevancy data. To increase diversity during training, random task instructions are generated with Claude 3 Sonnet based on hand-crafted templates. The demonstration examples are sampled from training data of the same relevancy task. The task instruction together with demonstration examples form the prompt input for MLLMs. It can vary across different samples.

\subsection{Binary Relevancy Data Creation}
As there are no available complex binary relevancy datasets for training and evaluation, we create data from diverse public datasets listed in Tab.~\ref{tab:list_of_datasets}. These are for preliminary experiments and we plan to expand them in a future work. The datasets will be released upon approval.

\begin{figure}[!t]
    \centering
    \includegraphics[width=1.0\columnwidth]{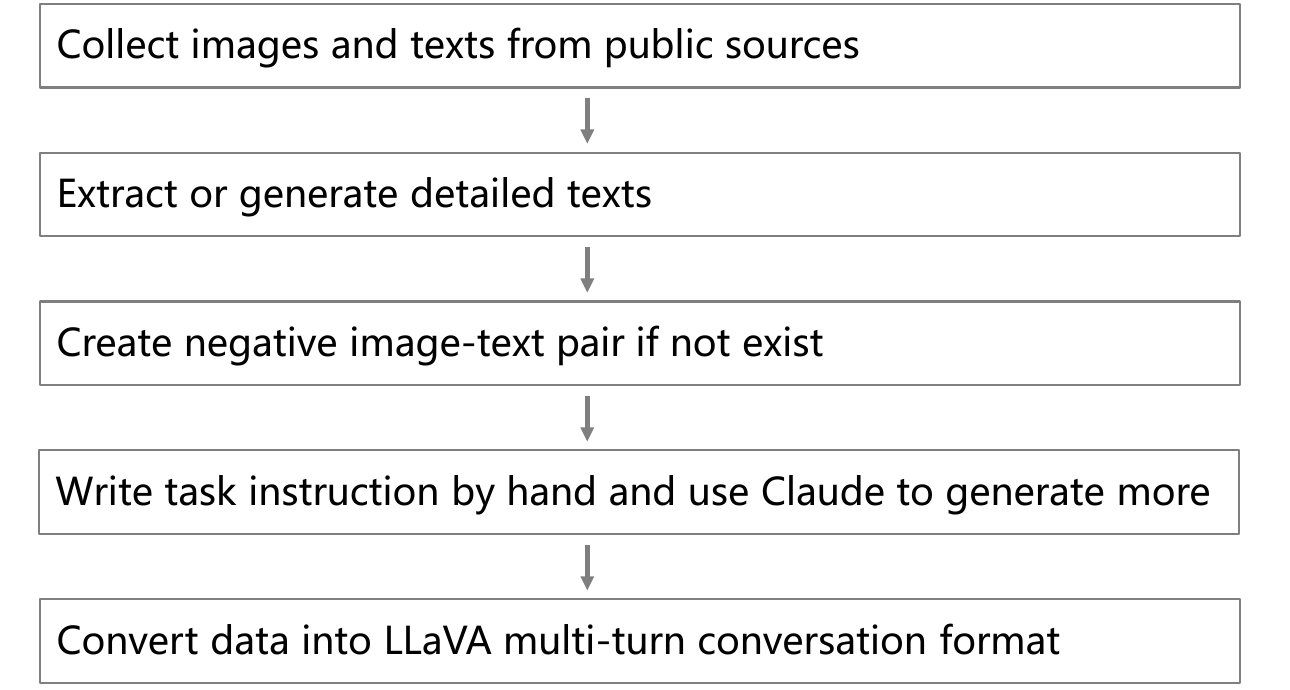}
    \caption{Data creation pipeline.}
    \label{fig:pipeline}
\end{figure}

\begin{figure*}[!t]
    \centering
    \includegraphics[width=0.8\textwidth]{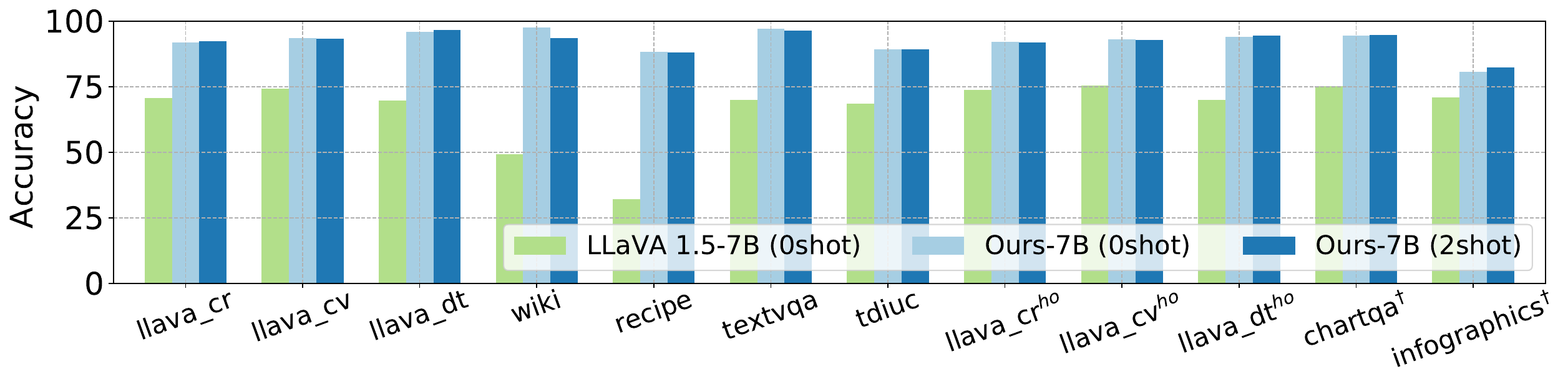}
    \includegraphics[width=0.8\textwidth]{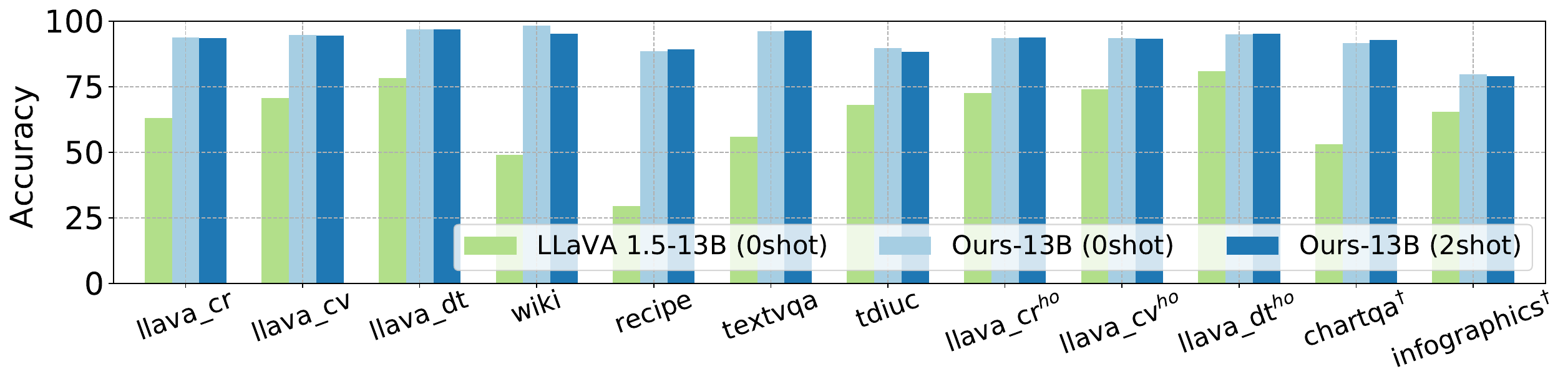}
    %\vspace{-4mm}
    \caption{Evaluation results on training and unseen tasks. ($^{ho}$hold-out test data, $^{\dagger}$unseen test tasks)}
    \label{fig:eval_train_unseen}
     % \vspace{-4mm}
\end{figure*}

The data creation pipeline consists of 5 stages, as shown in Fig.~\ref{fig:pipeline}. We first collect public data with image and text correspondences. As we focus on texts with complex formats and rich details, we select VQA datasets whose questions require some reasoning, and structured data like Wikipedia pages. Having more diverse data sources would certainly be helpful. The images are ready to use, while texts need additional processing. We extract related texts from the raw data, and format them with predefined templates. For short texts, we use Claude 3 Sonnet to generate detailed reasoning words or descriptions. After that, a key step is to define positive and negative image-text pairs. Positive pairs are easy to obtain as they can be derived from the raw data correspondences, while negative pairs may not exist. We create negative pairs by sampling images (or texts) from the same category. The specific strategies depend on the datasets. It is worth mentioning that defining proper negative pairs is a challenging task, as there are no human annotations. If using more strict rules (\textit{e.g.}, higher similarity score thresholds), some relevant image-text pairs may be mislabeled as `hard' negative samples. Finally, we use Claude 3 Sonnet to generate task instructions for each dataset and convert data into LLaVA multi-turn conversation format. Due to page length limit, more details can be found in Sec.~\ref{sec:app:br}.

%For model evaluation, we split test sets from the 6 datasets, and keep two hold-out tasks whose image categories are disjoint with training tasks. In addition, we create two more evaluation-only datasets from ChartQA~\cite{masry2022chartqa} and InfographicsVQA~\cite{mathew2022infographicvqa}. To further evaluate fine-grained relevancy, we create 6 tasks from commonly used fine-grained classification datasets including cars~\cite{krause20133d}, CUB~\cite{wah2011caltech}, dogs~\cite{dataset2011novel}, pets~\cite{parkhi2012cats}, flowers~\cite{nilsback2008automated} and food~\cite{bossard2014food}.

%Each dataset is divided into training and test splits. In total, there are 100$k$ training data. \texttt{llava} has 3 subtasks (cr, cv and dt). Meanwhile, we take \texttt{person} category as hold-out tasks that only appear in test data.

% For \texttt{news} and keep two hold-out tasks whose image categories are disjoint with training tasks. 

\subsection{Framework}

The framework of LLaVA-RE model is plotted in Fig.~\ref{fig:app:framework}. Given a pair of image and text, and a task instruction, we use ViT model~\cite{dosovitskiy2020image} to extract image tokens. The text and instruction are transformed using a default tokenizor. An MLP module maps original image tokens into text space. In addition, there are several context samples tokenized in the same way. The Vicuna model~\cite{vicuna2023} takes the entire token sequence and predicts a binary relevancy label.

\section{Experiments}
\subsection{Setup}
\paragraph{Model Settings.} Following LLaVA 1.5, we use CLIP-Large of 336$\times$336 image resolution as vision encoder and Vicuna as LLM backbone. We experiment with both 7B and 13B Vicuna models. 

\paragraph{Training Details.} We use the same pretraining setting as LLaVA 1.5 to learn the image projector. During instruction tuning phase, we conduct 4-shot ICL tuning. The 4 context samples are randomly selected from the same task. For each ICL training sample, losses are applied to both context samples and training sample. Thus, the effective training shot ranges from 0 to 4. We also include LLaVA-Instruct-665k into training to preserve general VQA capability, but only train with 0-shot. While the goal is to do relevancy evaluation, we find that ICL training with only relevancy data is prone to overfitting. To alleviate this, we add 24$k$ ICL samples of general VQA tasks created from TDIUC. The training is conducted with LoRA using 8$\times$ A100s. Input token length limit is set to 4096. As one image takes 576 tokens, it allows for 4 context samples in most cases. The learning rate of instruction tuning phase is 1.5e-4. Other hyper-parameters are kept the same as LLaVA 1.5.

\paragraph{Evaluation Details.} We evaluate binary prediction accuracies on the test split of training tasks, hold-out and unseen tasks. During inference, each dataset uses a hand-crafted task instruction that is unseen during training. The in-context samples are sampled from training data in a balanced manner, \textit{i.e.}, relevant and not relevant samples alternatively. In some rare cases where 4-shot inference exceeds 4096 tokens, we adjust token limit to 5120 to achieve a valid prediction.

%\paragraph{Comparison Methods.}
%We compare with vanilla LLaVA 1.5 models with 7B and 13B Vicunna backbones. Besides, Idefics2-8B~\cite{laurenccon2024matters} is a powerful vision-language model recently released. It was trained on a huge collection of image-text datasets. Its VQA performance competes with much larger models such as LLaVA-Next-34B~\cite{liu2024llavanext}. We use Idefics2 as a reference of strong open-source MLLMs, and use Claude3 for commercial MLLMs. 

\subsection{Results}

\begin{figure}[t]
    \centering
    \includegraphics[width=0.8\linewidth]{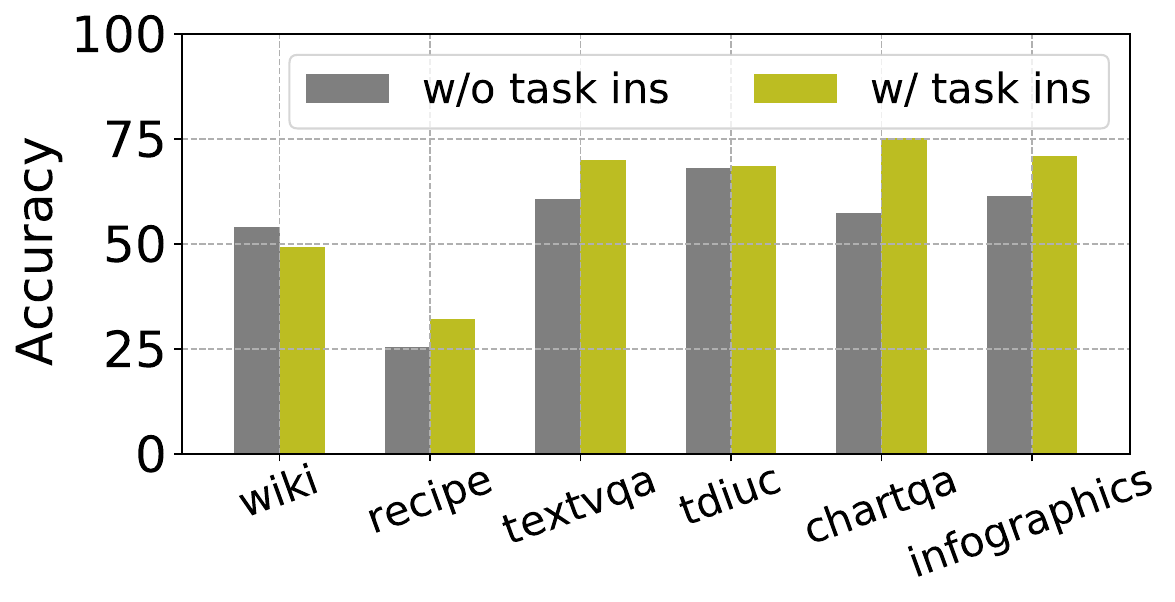}
    %\vspace{-2mm}
    \caption{Effect of task instructions on LLaVA 1.5-7B.}
    \label{fig:eval_task_ins}
     %\vspace{-2mm}
\end{figure}

\paragraph{Effect of Task Instructions.} We first study the effect of task instructions during inference on LLaVA 1.5 model. From Fig.~\ref{fig:eval_task_ins}, it can be seen that using task instructions achieves better accuracies on 5 out of 6 tasks. Since LLaVA 1.5 is not trained specifically for our binary image-text relevancy task, relevancy instructions provide useful information. 

\paragraph{Evaluation on Training and Unseen Tasks.}
Figure~\ref{fig:eval_train_unseen} plots the evaluation results of LLaVA 1.5 and our LLaVA-RE model. On the test split of 5 training tasks, LLaVA-RE achieves much higher accuracies than LLaVA 1.5 with both 7B and 13B Vicuna backbones. LLaVA 1.5's accuracies are below 50\% on some challenging tasks such as \texttt{wiki} and \texttt{recipe}, showing that binary image-text relevancy evaluation can sometimes be hard for off-the-shelf state-of-the-art models. The improvement of our models is consistent on 3 hold-out training tasks and 2 unseen tasks, which validates the generalization capability of LLaVA-RE. In the evaluations, 2-shot inference does not show much difference compared to 0-shot. One reason is that the ICL instruction tuning also optimizes 0-shot loss on the training tasks. Another reason is that 2-shot context samples are randomly selected and not semantically related to the test example.  

\begin{figure}[!t]
    \centering
    \includegraphics[width=\columnwidth]{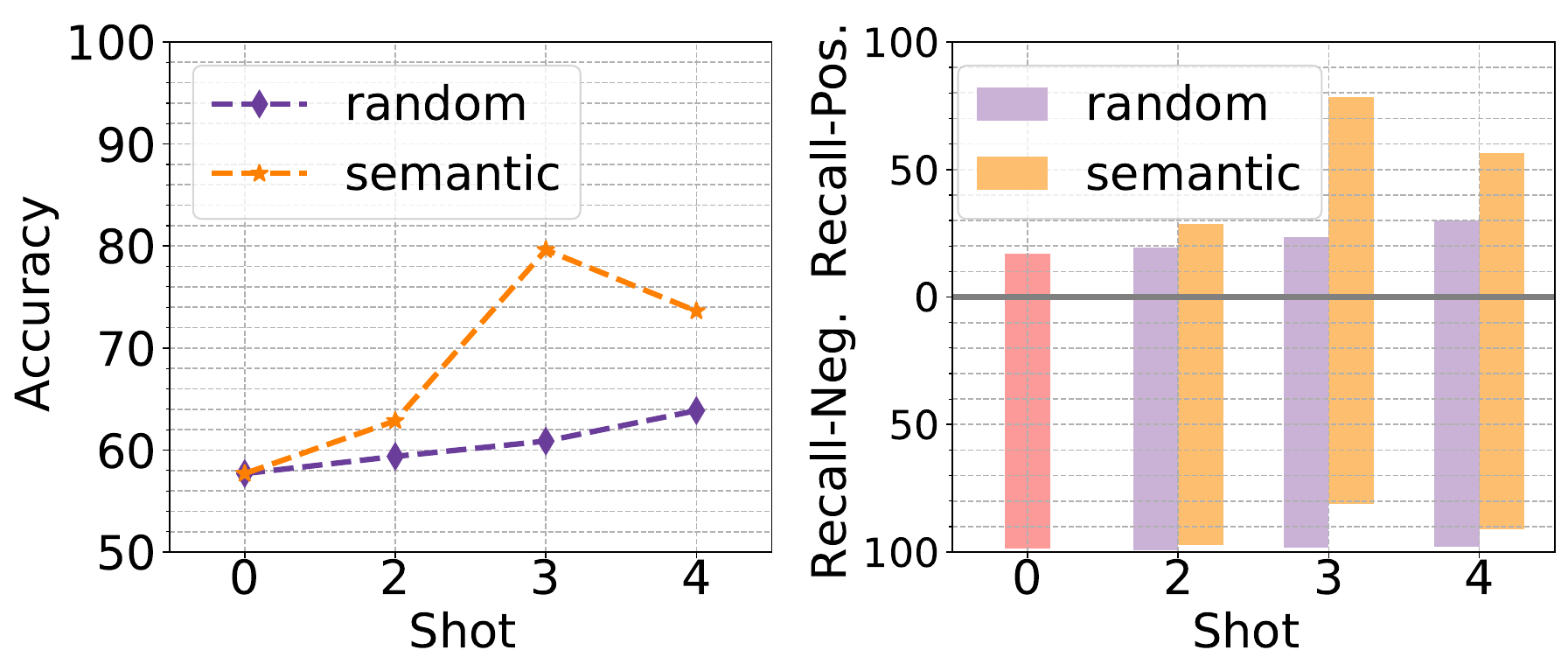}
    %\vspace{-6mm}
    \caption{Evaluation results averaged over 6 fine-grained tasks on Ours-7B, using random or sematic-related ICL context examples, (left): accuracies, (right): recalls for negative and positive samples. 0-shot results are shown in a red bar for a comparison.}
    \label{fig:eval_fine_grained}
    %\vspace{-2mm}
\end{figure}

\paragraph{Evaluation on Fine-grained Tasks.}
To further study the influence of ICL context examples, we evaluate LLaVA-RE on 6 fine-grained tasks. It is worth mentioning that these tasks are very different from the training and unseen test tasks in the previous subsection. The fine-grained classes have subtle definition and merely overlap with our training data. The left part of Fig.~\ref{fig:eval_fine_grained} plots the averaged accuracies under different numbers of shots. The ICL contexts are either random or semantic-related. In the former situation, context examples are randomly sampled from the whole dataset; in the latter situation, context examples share the same texts as the test example. The accuracy for 0-shot inference is unsatisfactory. From the recall plots on the right part of Fig.~\ref{fig:eval_fine_grained}, we see that the predictions are biased towards a negative answer (\textit{i.e.}, ``Not Relevant''). This could be attributed to the distribution shift between training and fine-grained tasks. When doing evaluation with ICL contexts (Shot>0), the predictions become more balanced and the overall accuracies improve over 0-shot results. Using semantic-related contexts clearly outperforms random contexts. These observations validate the effectiveness of ICL in our image-text relevancy evaluation.

\section{Conclusion}
In this paper, we study the important task of binary image-text relevancy evaluation. We present LLaVA-RE, a first attempt based on MLLM. It leverages task instructions and multimodal in-context samples to handle complex relevancy tasks. Furthermore, we create a novel binary relevancy dataset for training and evaluation. Experimental results validate the effectiveness of our framework. In future work, we plan to compare our model with more MLLMs and traditional semantic embedding models.

\section*{Limitations}

\paragraph{Relevancy task instructions.} This paper studies the evaluation of relevancy between image and complex text. In some scenarios, the definition of relevancy can be ambiguous if we focus on different aspects. For example, an image of a husky and a text of corgi can be regarded as relevant in terms of the general dog category, but irrelevant if focusing on the dog breeds. We use a few sentences of task instructions as the model input. However, how well an MLLM can follow those fine-grained instructions relies on the foundational capability of the LLM backbone. In NLP, an LLM foundation model usually requires a sufficient model size (\textit{i.e.} 13B) to have a good understanding of complex texts. Our training process does not own control of this instruction following capability. It will be necessary to understand how different LLM model sizes affect the understanding of easy and challenging relevancy tasks. Besides, the task instructions we use are written by human. It is helpful to explore the best form of task instructions for MLLM in our relevancy evaluation situation.

\paragraph{Context samples size.} 
Due to the 4096 input tokens limitation of LLM backbones, we can only use up to four context samples (image-text pairs). However, four samples may not be enough for some ambiguous relevancy tasks. One image takes 576 tokens, which has a large redundancy. There are some works showing that the number of image tokens can be even reduced from 576 to 9 without affecting the performance much~\cite{cai2024matryoshka}. In future work, it is meaningful to study how to combine these techniques into LLaVA-RE to incorporate more context samples.

\paragraph{Label noises.} We constructed multiple binary relevancy tasks from existing public datasets. The most challenging part is how to define negative pairs. In this paper, we use some heuristic ideas, such as sampling another image from the same category based on image similarities. However, the image similarity scores may not necessarily reflect the true fine-grained correlations, and this procedure inevitably introduces some noisy labels. Existing public multi-modal datasets mostly are not built to evaluate relevancy. How to construct high-quality relevancy labels, \textit{e.g.}, by expert annotations, is a challenging yet important problem for our future explorations.

\bibliography{reference}

\clearpage
\appendix

\renewcommand{\thefigure}{A.\arabic{figure}}
\setcounter{figure}{0}
\renewcommand{\thetable}{A.\arabic{table}}
\setcounter{table}{0}
\renewcommand{\theequation}{A.\arabic{equation}}
\setcounter{figure}{0}

\section{Appendix}
\label{sec:appendix}

\subsection{Data Creation Details}
\label{sec:app:br}

We created several binary relevancy datasets for training and evaluation, based on public data sources. Below we present creation details for each task.

\paragraph{LLaVA.}
LLaVA-Visual-Instruct 150K~\cite{liu2024improved} is constructed for visual instruction tuning by prompting GPT-4 API. It contains three subtasks: detailed description, conversation and complex reasoning. The images are from COCO dataset~\cite{lin2014microsoft}. Each raw sample contains one image $I$ and a series of questions and answers $\{(Q_i,A_i)\}$ related to the image. We convert the QA series into a long text $T$ by applying a simple template of ``Question: \{$Q$\} Answer: \{$A$\}'' on each QA. $(I,T)$ thus defines a positive image-text pair. To create non-relevant data, we randomly sample another image $\tilde{I}$ that belongs to the same category as $I$ and define $\{\tilde{I}, T\}$ as the negative pair. For evaluation purpose, we create hold-out test tasks using COCO \texttt{person} category, which is disjoint with the training categories.

\paragraph{Wiki.}
Wikipedia-based Image Text (WIT) Dataset~\cite{srinivasan2021wit} is a large multimodal multilingual dataset extracted from Wikipedia pages. The original data is composed of 37.6 million entity rich image-text examples across 108 Wikipedia languages, while we only use a very small portion of English sources (20k). One WIT data entry includes several fields such as \texttt{page title}, \texttt{page description}, \texttt{section text}, \texttt{section image}, \textit{etc}. Since \texttt{section text} and \texttt{section image} co-exist in the page, it is reasonable to define them as the positive image-text pair. On the other hand, \texttt{page description} describes the same topic as \texttt{section text} yet has different details, we define \texttt{page description} and \texttt{section image} as the negative image-text pair.

\paragraph{Recipe.}
RecipeQA~\cite{yagcioglu2018recipeqa} is a challenging dataset for multimodal comprehension of cooking recipes. Each recipe consists of textual descriptions of several steps to cook a particular food, and among them the first step usually talks about ingredients. Along with the recipe is a choice list of one positive image about this food and three negative images about other food. We filter out recipes whose first step has a title of ``ingredients''. Then we define positive image-text pair as positive food image and textual description of the first step, and similarly negative image-text pair with negative food images.

\paragraph{TextVQA.}
TextVQA~\cite{singh2019towards} is a VQA dataset that requires models to read and reason about text in images to answer questions about them. For example, one question is ``what kind of mushrooms are being advertised?'', and the answer is ``breaded''. All questions and answers are short. To create a long text for our relevancy evaluation purpose, we send the image, question and answer to Claude 3 Sonnet and ask Claude to generate a few sentences to justify the answer, \textit{e.g.}, \textit{``The advertisement clearly states `Try Our Enchanting Breaded Mushrooms' at the bottom, directly referring to breaded mushrooms as the featured item being promoted. The image reinforces this by depicting large, breaded mushroom-like structures alongside a character from Alice in Wonderland's whimsical setting, playing on the `wonderland' theme mentioned. Therefore, based on the explicit text and visual context provided, the type of mushrooms being advertised are indeed breaded mushrooms.''}. Given the question $Q$, answer $A$ and reasoning $R$, we apply a template of ``Question: \{$Q$\} Answer: \{$A$\}. \{$R$\}'' to create a long text. A positive pair is an image and the corresponding question and answer. To get a negative pair, we randomly sampled another image from the same category based on image similarity scores.

\paragraph{TDIUC.}
Task Driven Image Understanding Challenge (TDIUC)~\cite{kafle2017analysis} is a VQA dataset organized into 12 different categories. Each category focuses on a particular task such as \texttt{object presence}, \texttt{sport recognition}, \textit{etc}. It also introduces a category of \texttt{absurd} questions that are meaningless for a given image. To make binary relevancy tasks, we only use data from three categories: \texttt{activity recognition}, \texttt{sentiment understanding} and \texttt{utility/affordance} which require more reasoning capability. We use Claude 3 Sonnet to generate a few sentences of justification based on the image, question and answer. A positive pair is an image and the corresponding question and answer, using a similar template as \texttt{textvqa}. To get a negative pair, we randomly sample another question/answer from the same category based on text similarity scores.

\paragraph{ChartQA.}
ChartQA~\cite{masry2022chartqa} is a benchmark for question answering about chart images. These images are different from training tasks, and used to evaluate models' generalization ability. We use Claude 3 Sonnet to generate a few sentences to justify the answer, and create positive/negative pairs in a similar manner as \texttt{textvqa}. 

\paragraph{Infographics.}
InfographicVQA~\cite{mathew2022infographicvqa} comprises a diverse collection of infographics with question-answer annotations. The questions require elementary reasoning and basic arithmetic skills over the document layout, textual content, graphical elements and data visualization. We use Claude 3 Sonnet to generate a few sentences to justify the answer, and create positive/negative pairs in a similar manner as \texttt{tdiuc}.

\paragraph{Fine-grained.} 
 We create 6 tasks from commonly used fine-grained classification datasets including cars~\cite{krause20133d}, CUB~\cite{wah2011caltech}, dogs~\cite{dataset2011novel}, pets~\cite{parkhi2012cats}, flowers~\cite{nilsback2008automated} and food~\cite{bossard2014food}. For each dataset, we ask Claude 3 Sonnet to generate useful visual features to distinguish one class. With this, each class label is converted into a long textual description focusing on fine-grained visual features. For example, give a car model ``Dodge Caliber Wagon 2007'', Claude responses with \textit{``The Dodge Caliber Wagon 2007 has a distinctive boxy and upright shape with a tall stance and pronounced wheel arches. Its front end features a characteristic crosshair grille with the Dodge logo in the center, and angular headlights that sweep back towards the fenders. The side profile shows a long greenhouse with an upswept beltline and a rear quarter window, giving it a distinctive wagon silhouette. The wheels are typically five-spoke alloy rims, and the body color options range from bold shades like Infrared and Sunburst Orange to more subdued hues like Silver and Black. Distinctive badging on the rear liftgate and lower body panels proudly displays the `Caliber' name and Dodge branding.''} The positive pair is an image and the corresponding class description. To get a negative pair, we randomly sample another image from the same fine-grained class.

\begin{table*}[!t]
	\caption{Detailed evaluation accuracies on training and unseen tasks. (`lv' short for `llava', `info.' short for infographics, $^{ho}$hold-out test data, $^{\dagger}$unseen test tasks)}
	\footnotesize
	\centering
        \renewcommand{\tabcolsep}{0.1cm}
	\scalebox{1.0}{
		\begin{tabular}{p{2cm}p{0.5cm}<{\centering}p{0.85cm}<{\centering}p{0.85cm}<{\centering}p{0.85cm}<{\centering}p{0.85cm}<{\centering}p{0.85cm}<{\centering}p{0.85cm}<{\centering}p{0.85cm}<{\centering}p{0.85cm}<{\centering}p{0.85cm}<{\centering}p{0.85cm}<{\centering}p{0.85cm}<{\centering}p{0.85cm}<{\centering}}
			\toprule
			model & shot & lv\_cr & lv\_cv & lv\_dt & wiki & recipe & textvqa & tdiuc & lv\_cr$^{ho}$ & lv\_cv$^{ho}$ & lv\_dt$^{ho}$ &  chartqa$^{\dagger}$ & info.$^{\dagger}$\\ 
			\midrule
 LLaVA 1.5-7B & 0 & 70.8 & 74.3 & 69.7 & 49.3 & 32.2 & 70.0 & 68.7 & 73.9 & 75.4 & 70.0 & 75.1 & 70.9 \\
 Ours-7B & 0  & 91.9 & 93.5 & 96.0 & 97.7 & 88.3 & 97.1 & 89.3 & 92.2 & 93.1 & 94.1 & 94.6 & 80.8 \\
 Ours-7B & 2  & 92.4 & 93.4 & 96.7 & 93.7 & 88.2 & 96.5 & 89.3 & 92.0 & 92.8 & 94.5 & 94.8 & 82.3 \\
\midrule
  LLaVA 1.5-13B & 0  & 63.0 & 70.6 & 78.4 & 49.0 & 29.6 & 55.9 & 68.0 & 72.5 & 74.0 & 80.9 & 53.1 & 65.4 \\
  Ours-13B & 0  & 93.7 & 94.7 & 97.0 & 98.3 & 88.6 & 96.3 & 89.7 & 93.6 & 93.6 & 95.0 & 91.8 & 79.7 \\
  Ours-13B & 2  & 93.6 & 94.6 & 97.0 & 95.3 & 89.3 & 96.4 & 88.3 & 93.7 & 93.3 & 95.2 & 93.0 & 79.1 \\
			\bottomrule
	\end{tabular} }
	\label{tab:app:results_train_unseen}
	% \vspace{-1mm}
\end{table*}

\subsection{Detailed Accuracies}
Table~\ref{tab:app:results_train_unseen} lists the detailed accuracies of comparison models on training and unseen tasks.

\subsection{Sampled Task Instructions}
Below we show some task instructions from sampled datasets. 

\paragraph{LLaVA.}
\textit{You are given an image and a series of Question and Answer. Your task is to judge whether the image is relevant to these Questions and Answer. Here are several important instructions:\\
- Do not simply confirm the the object exists in image.\\
- Think about whether there is visual evidence supports or unrelated or contradicts the question and answer.\\
- In the textual question and answer, look for attributes such as color, size, shape, location, etc. And evaluate if the image matches these attributes.\\
- In the textual question and answer, look for context or settings of how the object is shown (background, neighboring objects, usage scenarios, etc.), and evaluate if the image shows the context.\\
- Use only the clear visual information that can be directly seen from image to determine the relevancy to question and answers.\\
- IMPORTANT: do not reason with your own knowledge or additional hallucination or guessing to determine relevancy.\\
- IMPORTANT: do not say `yes' if certain aspects cannot be determined visually, Look very careful at the image!\\
- IMPORTANT: do not say `yes' if answering requires knowledge beyond the image.\\
- Only say `yes' if the image shows direct and obvious matching visual clues that supports the textual question and answer.\\
- If there are multiple question and answer, only say `yes' if the image is relevant to all question and answer.\\
- If image is only related to the object and does not match the attributes, you should say `no'.}

\paragraph{Textvqa.} 
\textit{You are given an image and a pair of question and answer. Your task is to judge whether the image is relevant to the question and answer. Here are several important instructions:\\
- The question focuses on text understanding. The image may be coherent or incoherent to this question.\\
- The answer includes an explanation to justify itself. It contains important details about a true relevant image.\\
- In the text, look for descriptions about objects, characters, colors, spatial relationships. Check whether these descriptions match the image.\\
- In the image, recognize existing characters such as digits, english letters, before making a judgement.\\
- Use only the clear visual information that can be directly seen from image to determine the relevancy to text.\\
- IMPORTANT: do not reason with additional hallucination or guessing to determine relevancy.\\
- IMPORTANT: do not say `yes' if certain aspects cannot be determined visually, Look very careful at the image!\\
- Only say `yes' if the image shows direct and obvious matching visual clues that supports the text.\\
- If image contradicts with answer regarding the question, you should say `no'.\\
- The answer must be a single word of `Yes' or `No'.}

\paragraph{Cars (fine-grained).} 
\textit{You are given a car image and a short description about a specific car model. Your task is to judge whether the image is relevant to the text. Here are several important instructions:\\
- Carefully look at details in the image, such as car shape, decoration, color, number of doors, wheel sizes.\\
- The image may look similar to the described car model, but not exactly match it.\\
- Use your own knowledge to distinguish any visual differences between the image and the car description.\\
- Only say `yes' if the image shows exactly the same fine-grained attributes as the description. Otherwise, say `no'.\\
- The answer must be a single word of `yes' or `no'.}

\subsection{Sampled Image-Text pairs}
We show some sampled image-text pairs from training and test tasks in Tables~\ref{tab:sample:it_pairs:I},\ref{tab:sample:it_pairs:II}. 

\begin{table*}
\caption{Sampled image-text pairs from different tasks.} \label{tab:sample:it_pairs:I}
\centering
\footnotesize
\begin{tabular}{p{0.3\textwidth}p{0.6\textwidth}}
\toprule

\begin{minipage}{0.3\textwidth}
Task: llava; Label: `relevant' \\
\includegraphics[width=\textwidth]{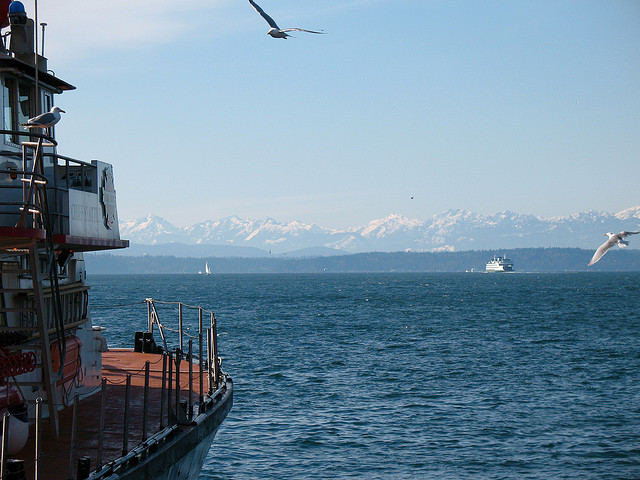}
\end{minipage} &  \begin{minipage}{0.6\textwidth} 
Question: What role might a ferry play in a location like this?

Answer: In this oceanic scene with a view of the mountains, a ferry might play a significant role in providing transportation services between various coastal points, including towns, cities, and islands. As the landscape is surrounded by a large body of water, the ferry connects these locations and enables people to travel across the water efficiently and conveniently. It can be a preferred mode of transport for commuters, tourists, or locals who need to access services or visit attractions in different areas. Additionally, ferries might also accommodate vehicles and cargo, providing further convenience to travelers and playing a crucial role in the region's economy by facilitating trade and the movement of goods.

Is the image relevant to this Question and Answer?
\end{minipage} \\

\midrule

\begin{minipage}{0.3\textwidth}
Task: llava; Label: `not relevant' \\
\includegraphics[width=\textwidth]{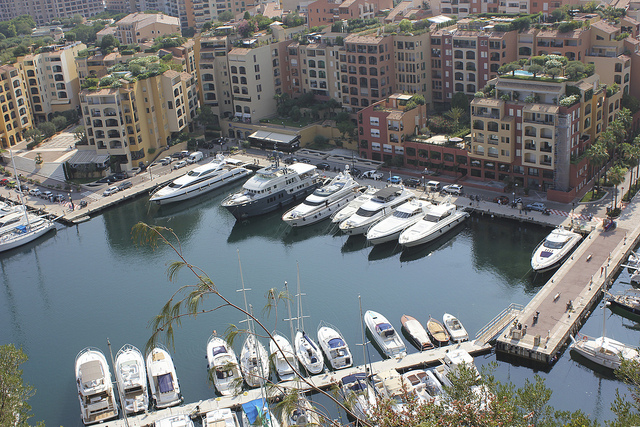}
\end{minipage} &  \begin{minipage}{0.6\textwidth}
Question: Explain the visual content of the image in great detail.

Answer: The image captures a city street scene at either dusk or dawn with a flock of birds flying high above. There are at least 15 birds scattered across the sky in various positions, creating a sense of motion and activity. 

Along the street, cars are parked on the side, and some are in motion, with one of the moving cars having its headlights on. The street itself is lined with trees and buildings, adding to the urban atmosphere.

Traffic lights are also visible in the scene, with three on the left side and two on the right side of the street. The overall mood of the image is serene as the sun goes down, creating an end-of-the-day feeling in the city.

Is the image relevant to this Question and Answer?
\end{minipage} \\

\midrule

\begin{minipage}{0.3\textwidth}
Task: wiki; Label: `relevant' \\
\includegraphics[width=\textwidth]{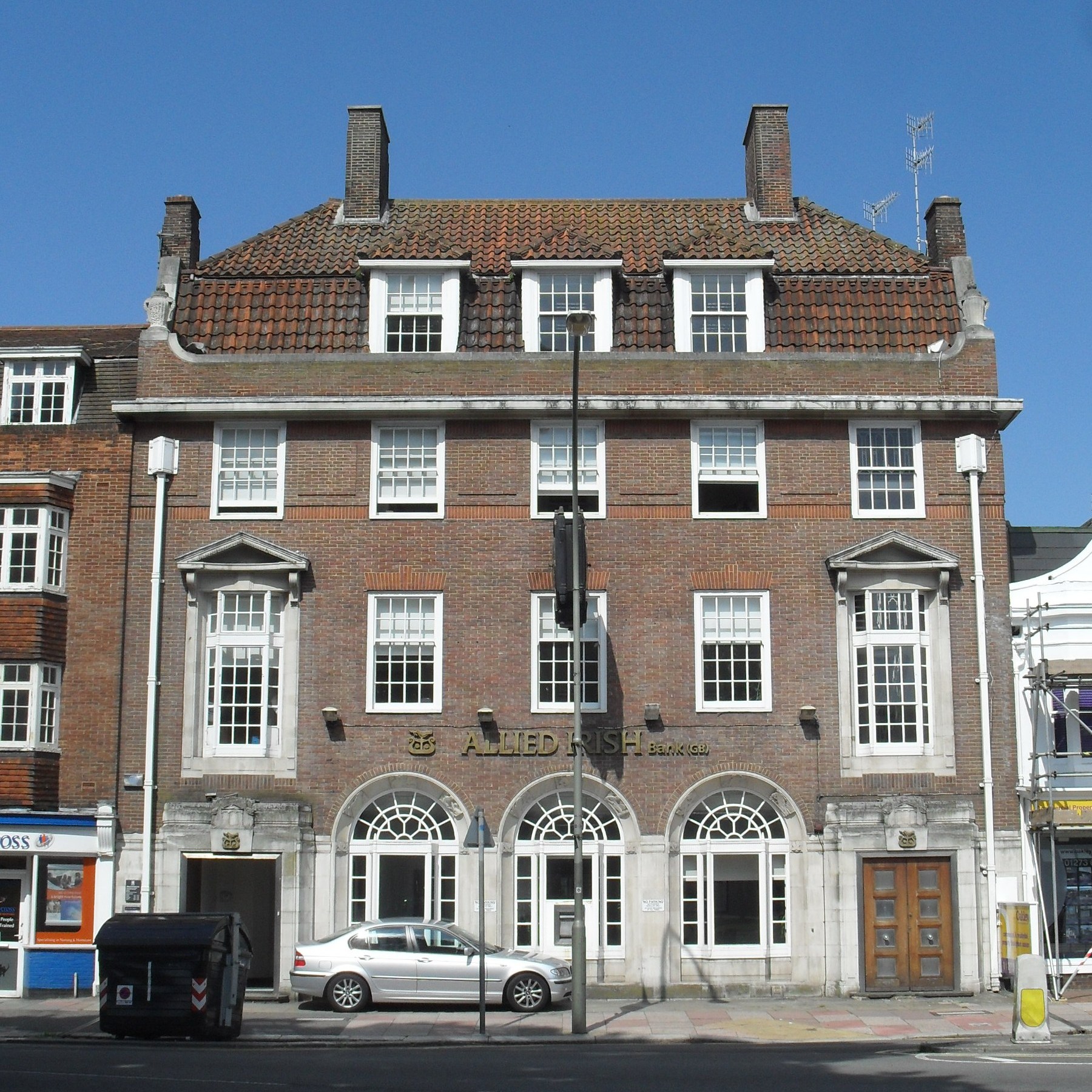}
\end{minipage} &  \begin{minipage}{0.6\textwidth}
Text: A pantile is a type of fired roof tile, normally made from clay.  It is S-shaped in profile and is single lap, meaning that the end of the tile laps only the course immediately below.  Flat tiles normally lap two courses.

A pantile-covered roof is considerably lighter than a flat-tiled equivalent and can be laid to a lower pitch.

Pantiles are used in eastern coastal parts of England and Scotland including Norfolk, East Yorkshire, County Durham, Perthshire, Angus, Lothian and Fife, where they were first imported from the Netherlands in the early 17th century. They are rarely used in western England or western Scotland, except in Bristol and the Somerset town of Bridgwater.

Roofing pantiles are not to be confused with a type used for paving, after which the Georgian colonnade in Tunbridge Wells is named.  Whilst called pantiles, the paving tiles which were installed there in 1699 were one-inch-thick square tiles made from heavy wealden clay, so-named as shaped in a wooden pan before firing. The pantile paving in Tunbridge Wells was replaced with stone flag tiles in 1792.

Is the image relevant to this Text?

\end{minipage} \\

\midrule

\begin{minipage}{0.3\textwidth}
Task: recipe; Label: `relevant' \\
\includegraphics[width=\textwidth]{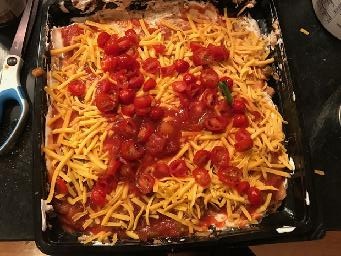}
\end{minipage} &  \begin{minipage}{0.6\textwidth} 

Ingredients: First of all collect the following ingredients (this is what I used for mine but if you want bigger or smaller meat loaf use your own quantity): 700g Minced meat 2 Packages of bacon 4 eggs (1 raw, 3 boiled) Ham Salt Pepper Parmigiano Cheese

Are the Ingredients necessary to make food in the image?

\end{minipage} \\

\midrule

\begin{minipage}{0.3\textwidth}
Task: textvqa; Label: `not relevant' \\
\includegraphics[width=\textwidth]{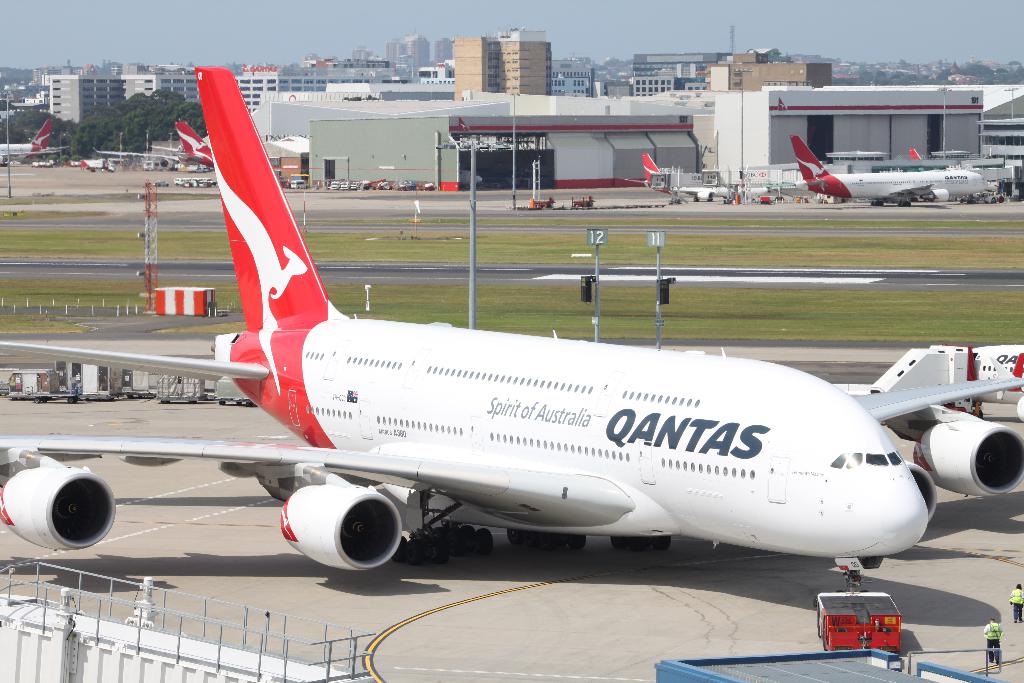}
\end{minipage} &  \begin{minipage}{0.6\textwidth} 
Question: what type of plane is this?

Answer: lape. The image depicts an old propeller-driven aircraft sitting on a grassy field. The registration or name painted on the side of the aircraft fuselage clearly reads "EC-AGC LAPE", indicating that the type of plane shown is a Lape model. The black and white photograph captures this vintage aircraft in an outdoor setting, providing visual evidence that confirms the stated answer that this is indeed a Lape type of plane.

Is the image relevant to this Question and Answer?
\end{minipage} \\

\bottomrule
\end{tabular}
\end{table*}

\begin{table*}
\caption{Sampled image-text pairs from different tasks.}\label{tab:sample:it_pairs:II}
\centering
\footnotesize
\begin{tabular}{p{0.3\textwidth}p{0.6\textwidth}}
\toprule

\begin{minipage}{0.3\textwidth}
Task: tdiuc; Label: `relevant' \\
\includegraphics[width=\textwidth]{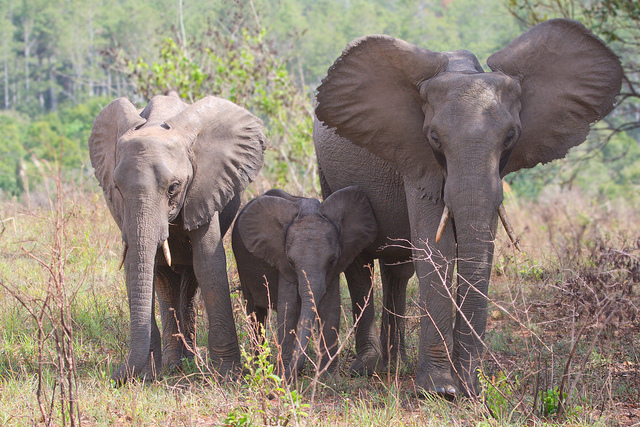}
\end{minipage} &  \begin{minipage}{0.6\textwidth} 
Question: What are elephants thought to be afraid of?

Answer: mice. While the image shows a group of elephants calmly walking together, it does not directly relate to what elephants may be afraid of. However, the provided answer that elephants are thought to be afraid of mice is an interesting folk belief or myth.

This notion likely stems from observations or anecdotes of elephants appearing startled or reacting to the sudden movements or sounds of small rodents, which could seem disproportionate to their massive size. However, scientific evidence suggests that elephants' fear of mice is likely exaggerated, and they may simply be reacting to unexpected stimuli rather than being genuinely afraid of the mice themselves.

While the origins of this belief are unclear, it persists as an enduring myth or legend about these majestic creatures. The image alone cannot confirm or deny this claim, as it simply depicts elephants in their natural habitat without any mice present.

Is the Answer relevant to this Image and Question?
\end{minipage} \\

\midrule

\begin{minipage}{0.3\textwidth}
Task: infographics; Label: `not relevant' \\
\includegraphics[width=\textwidth]{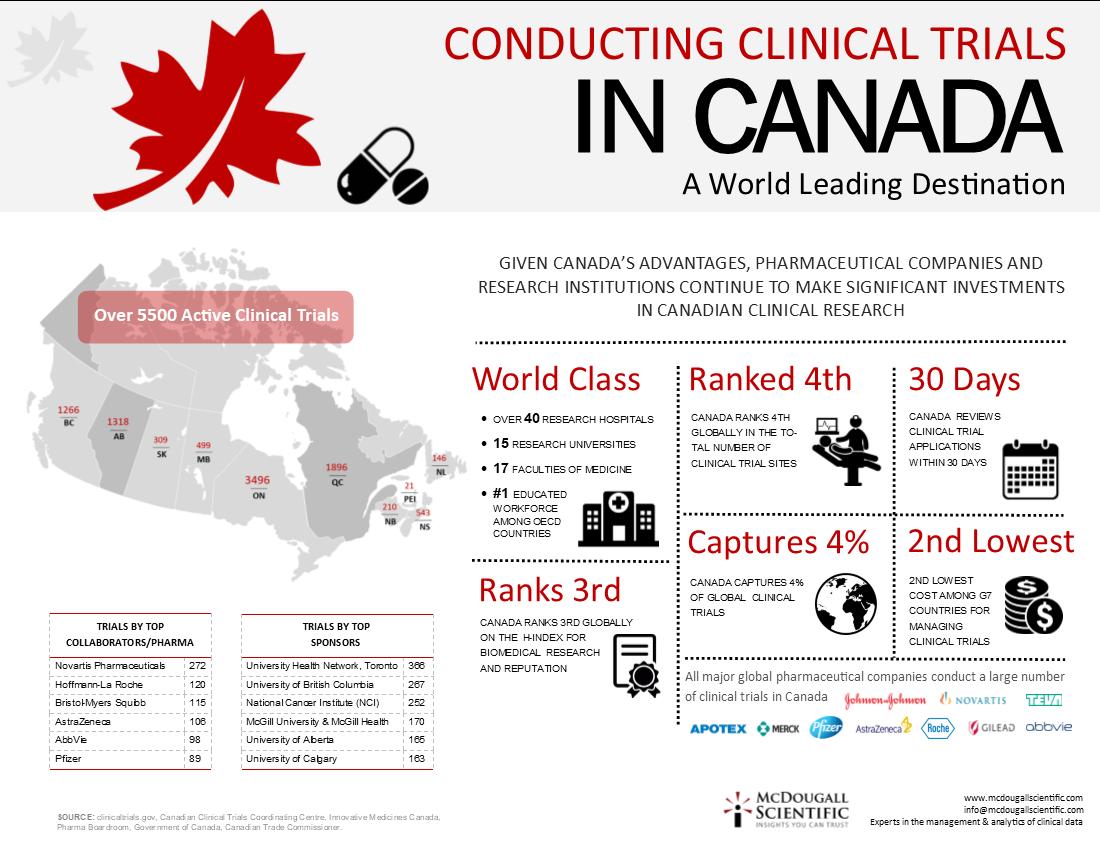}
\end{minipage} &  \begin{minipage}{0.6\textwidth} 
Question: What is the majority percentage of females in the wholesale \& retail industry of Canada? 

Answer: 21.0\%

Question: What percent of the women entrepreneurs in Canada work for 31-39 hours a week? 

Answer: 19.8\%

Question: What percent of the women employees in Canada work for 41-49 hours a week? 

Answer: 2.7\%

Question: How many hours per week is spend on work by 17.4\% of the women entrepreneurs in Canada? 

Answer: 40 HOURS

Is the image relevant to these Question and Answer?

\end{minipage} \\

\midrule

\begin{minipage}{0.3\textwidth}
Task: chartqa; Label: `not relevant' \\
\includegraphics[width=\textwidth]{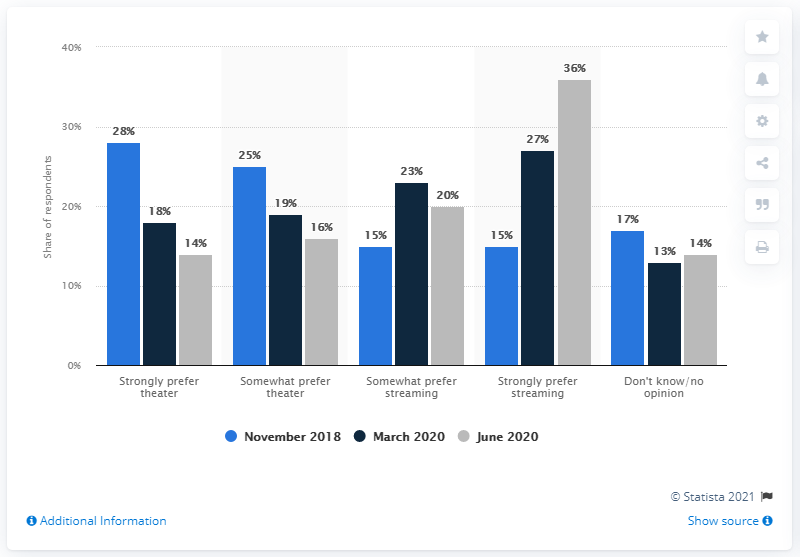}
\end{minipage} &  \begin{minipage}{0.6\textwidth} 
Question: What is the sum of making progress, losing ground and same in the year 2011?

Answer: 99. The image shows a line graph depicting the perceptions of Mexicans regarding the government's progress against drug trafficking over the years 2011 to 2015. In 2011, the values for "Making progress," "Losing ground," and "Same" add up to 45, 30, and 19 respectively, which totals 94. However, the provided answer of 99 does not correspond to the sum of these three values in 2011 based on the information given in the graph. To justify the answer, I would need additional context or clarification on the data being referenced.

Is the image relevant to this Question and Answer?

\end{minipage} \\

\midrule

\begin{minipage}{0.3\textwidth}
Task: flowers; Label: `relevant' \\
\includegraphics[width=\textwidth]{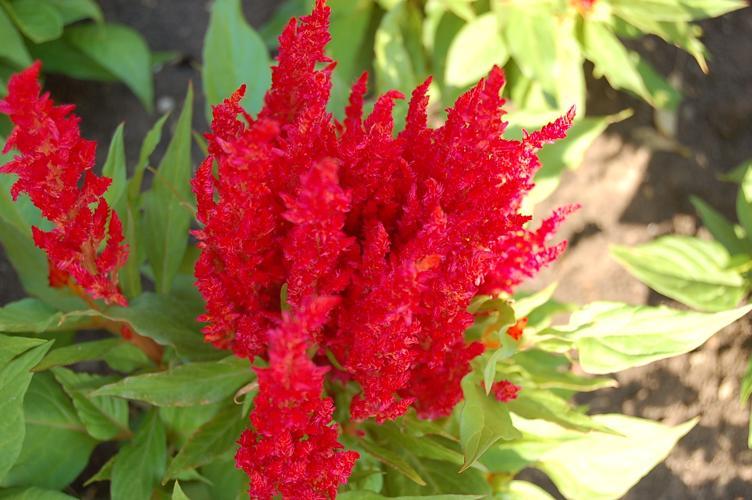}
\end{minipage} &  \begin{minipage}{0.6\textwidth}
The prince of wales feathers is a perennial flowering plant with tall spikes of red, velvety flowers. The vibrant red petals are tightly clustered together in a cylindrical shape, forming a distinctive feather-like appearance. The flowers emerge from a terminal spike, with overlapping bracts that provide a protective covering. The lance-shaped leaves are mid-green in color and arranged oppositely along the stem. The plant can grow up to 1.5 meters tall, with multiple flowering spikes emerging from a single stem. 

Is the image relevant to this flower description?
\end{minipage} \\

\midrule

\begin{minipage}{0.3\textwidth}
Task: cars; Label: `not relevant' \\
\includegraphics[width=\textwidth]{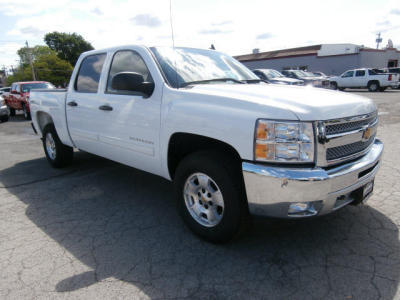}
\end{minipage} &  \begin{minipage}{0.6\textwidth}
The Ford Fiesta Sedan 2012 has a compact, three-box sedan body style with a distinct front grille featuring the iconic Ford blue oval logo in the center. Its headlights are swept back and have a distinctive shape, while the taillights have a distinctive LED light signature. The side profile features pronounced wheel arches and a character line running along the length of the car. The alloy wheels have a multi-spoke design and are typically 15 or 16 inches in diameter. Depending on the trim level, the exterior may feature body-colored door handles, side mirrors, and other accents, while higher trims may have chrome accents. 

Is the image relevant to this car description?
\end{minipage} \\

\bottomrule
\end{tabular}
\end{table*}

\end{document}